\documentclass{article}

\PassOptionsToPackage{numbers, compress}{natbib}
\usepackage[preprint]{neurips_2026}

\usepackage[utf8]{inputenc}
\usepackage[T1]{fontenc}
\usepackage{hyperref}
\usepackage{url}
\usepackage{booktabs}
\usepackage{amsfonts}
\usepackage{amsmath}
\usepackage{nicefrac}
\usepackage{microtype}
\usepackage{xcolor}
\usepackage{colortbl}
\usepackage{graphicx}
\usepackage{multirow}
\usepackage{enumitem}
\usepackage{pifont}
\usepackage{tcolorbox}
\usepackage{tikz}
\usepackage{soul}

\tcbuselibrary{skins,breakable}

\newcommand{\method}{\textsc{TokenMem}}

\newcommand{\vanillarag}{\textsc{VanillaRAG}}
\newcommand{\nomem}{\textsc{No-Memory}}

\newcommand{\gate}{W_\beta}
\newcommand{\gateA}{A_\beta}
\newcommand{\gateB}{B_\beta}
\newcommand{\hidden}{\mathbf{h}}
\newcommand{\query}{\mathbf{q}}
\newcommand{\key}{\mathbf{k}}

\newcommand{\mem}{\mathbf{m}}
\newcommand{\dhead}{d_h}
\newcommand{\nlayers}{L}

\newcommand{\cmark}{\textcolor{okGreen}{\ding{51}}}
\newcommand{\xmark}{\textcolor{gray!50}{\ding{55}}}

\newcommand{\pp}{\,\text{pp}}

\definecolor{tmBlue}{RGB}{41,98,255}
\definecolor{ragOrange}{RGB}{230,81,0}
\definecolor{okGreen}{RGB}{27,94,32}
\definecolor{badRed}{RGB}{183,28,28}
\definecolor{keyBg}{RGB}{228,237,252}

\definecolor{bestCell}{HTML}{E3F2FD}
\definecolor{oursRow}{HTML}{E3F2FD}
\newcommand{\best}[1]{\cellcolor{bestCell}\textbf{#1}}

\newtcolorbox{examplebox}[1][]{
  enhanced,
  colback=white,
  frame hidden,
  borderline={0.8pt}{0pt}{black!40,dashed},
  arc=4mm,
  left=8pt, right=8pt, top=3pt, bottom=3pt,
  boxsep=2pt,
  before skip=4pt,
  after skip=8pt,
  #1
}
\newcommand{\cfcorrect}[1]{\textcolor{okGreen}{\textbf{#1}}}
\newcommand{\cfwrong}[1]{\textcolor{badRed}{\textbf{#1}}}
\newcommand{\dashedline}{\noindent\tikz{\draw[dashed,black!30] (0,0) -- (\linewidth,0);}}
\sethlcolor{keyBg}
\newcommand{\keyphrase}[1]{\hl{#1}}
\newcommand{\methodbadge}[4]{%
  \tikz[baseline=(badge.base)]{%
    \node[fill=#1,draw=#2,rounded corners=3pt,
          inner xsep=5pt,inner ysep=2.5pt,
          font=\sffamily\scriptsize,line width=0.6pt] (badge)
      {\textcolor{#2}{\textbf{#3}}\ifx&#4&\else\textcolor{black!50}{$\bullet$}\hspace{2pt}#4\fi};%
  }%
}
\newcommand{\tmheader}[1]{\methodbadge{tmBlue!10}{tmBlue}{TokenMem}{#1}}
\newcommand{\ragheader}[1]{\methodbadge{ragOrange!10}{ragOrange}{VanillaRAG}{#1}}

\title{TokenMem: Faithful Knowledge Injection for Frozen LLMs}

\author{%
  Chengzhang Yu$^{1}$ \quad Chenyang Zheng$^{1}$ \quad Zening Lu$^{1}$ \quad Yingru He$^{1}$ \\
  Yutong Huang$^{1}$ \quad Yiming Zhang$^{2}$ \quad Yue Xu$^{1}$ \quad Zhanpeng Jin$^{1}$ \\
  \vspace{2pt} \\
  $^{1}$South China University of Technology \quad
  $^{2}$University of Science and Technology of China \\
}

\begin{document}

\maketitle

\begin{abstract}
Retrieval-augmented generation (RAG) enhances large language models (LLMs) with external knowledge, but suffers from \emph{knowledge conflicts}: when retrieved information contradicts parametric memory, the shared self-attention pathway produces unpredictable outputs.
We present \method{}, a lightweight memory system that injects knowledge into frozen LLMs through a dedicated cross-attention channel, bypassing competition with parametric memory in the residual stream.
\method{} trains only a thin gating adapter (${\sim}$3--7M parameters) via a two-phase curriculum: first learning general knowledge utilization, then strengthening faithful compliance under counterfactual knowledge.
In controlled experiments on five models spanning three families (Qwen3-4B/8B/14B, LLaMA-3.1-8B, OLMo-3-7B), \method{} achieves 69--70\% Knowledge Compliance (KC) on counterfactual benchmarks, compared to 20--52\% for vanilla RAG---a gap of up to 49 percentage points.
Ablation studies show that the two-phase curriculum is critical: removing Phase~2 collapses KC to near-zero. Mechanistic analysis reveals that the gate adapter learns a conflict-aware, layer-specific injection strategy without explicit supervision.
Code is available at \href{https://anonymous.4open.science/r/TokenMem0-D35B/README.md}{our anonymous repository}.
\end{abstract}

\section{Introduction}
\label{sec:intro}

The parametric knowledge of large language models (LLMs) is inevitably incomplete and becomes stale as the world evolves~\citep{kasai2023realtime, zhang2025evolvebench, lazaridou2021mind}.
In high-stakes settings such as financial compliance, clinical decision support, or legal advisory, relying on outdated parametric knowledge can produce harmful outputs~\citep{jin2025kpgap, luu2022time}.
Retrieval-augmented generation (RAG)~\citep{lewis2020rag} addresses this gap by supplying external evidence at inference time, with the implicit promise that models will \emph{faithfully ground} their outputs in the provided knowledge.
When the retrieved passage is designated as authoritative (e.g., a revised regulatory filing or an updated clinical guideline), the model should follow it even if its parametric prior disagrees.
We call this property \emph{Knowledge Compliance} (KC): given a passage the system treats as authoritative, KC measures how often the model follows it regardless of factual correctness---a \emph{controllability} metric analogous to honoring a database write, distinct from accuracy under noisy or adversarial retrieval.

Yet this promise breaks down precisely when it matters most.
When retrieved knowledge contradicts the model's parametric memory---the \emph{knowledge conflict} setting---LLMs exhibit erratic, unpredictable behavior~\citep{xie2024adaptive, zhao2024conflictbank}.
For instance, a model trained on pre-2024 medical data may ignore a retrieved passage stating a revised drug interaction, instead generating the outdated recommendation stored in its parameters.
\citet{xu2024kcsurvey} show that this is not an isolated failure: models exhibit a \emph{systematic bias} toward parametric memory, and larger models with stronger parametric priors show \emph{lower} compliance with external evidence, an inverse scaling phenomenon that undermines the foundational assumption of RAG.
Even when retrieval is accurate, the mere presence of additional context can paradoxically increase hallucination propensity rather than improve faithfulness~\citep{joren2025sufficient}.

We hypothesize that a key contributing factor is \emph{architectural}.
In standard RAG, external knowledge and query tokens share a single self-attention softmax: the model's parametric confidence competes with external evidence for a fixed attention budget.
When frozen weights produce strong key-query alignments among query tokens, attention to knowledge tokens is suppressed---one plausible mechanism behind why larger models are less compliant (\S\ref{sec:problem}).
This mechanism is \emph{architecturally addressable}: routing external knowledge through a separate softmax eliminates the competition by construction (Figure~\ref{fig:overview}).

This leads to our central hypothesis: \textbf{injecting external knowledge through an independent channel that does not share the attention budget with parametric memory should improve knowledge compliance under conflict.}
Cross-attention provides such a channel: it computes attention over external representations under its own normalization, leaving the self-attention pathway undisturbed.
We test this hypothesis by building a system around cross-attention injection and evaluating it across five models and three architectural families.

Based on this reasoning, we present \method{}, a lightweight memory system for frozen LLMs.
\method{} consists of three components: (1)~a \emph{TokenMemoryBank} that stores and retrieves knowledge passages via FAISS indexing; (2)~a \emph{cross-attention adapter} (adapted from DecoupledRAG~\citep{decoupledrag2025}) that injects retrieved knowledge through a zero-initialized gate at every layer of a frozen LLM; and (3)~a \emph{CoT curriculum SFT} protocol that trains the adapter (${\sim}$3--7M parameters) in two phases, first on knowledge utilization and then on compliance under conflict conditions.

High KC is the design goal for a controllable memory system: when the user writes a fact into the memory bank, the model should use it regardless of what its parameters encode.

Our key contributions are:

\begin{itemize}[leftmargin=*,itemsep=2pt]
    \item \textbf{System:} \method{}, a plug-and-play memory system for frozen LLMs that achieves high knowledge compliance through cross-attention injection. We validate across 5 models spanning 3 architectural families (Qwen3, LLaMA-3.1, OLMo-3).

    \item \textbf{Training protocol:} We show that a two-phase curriculum (utilization then compliance) is necessary: Phase~1 alone collapses KC to near-zero despite the cross-attention channel being present. This demonstrates that the channel provides \emph{capacity} for faithful injection, but explicit counterfactual training is required to activate it.

    \item \textbf{Empirical observations:} We observe inverse scaling in vanilla RAG's knowledge compliance---larger models show lower compliance, consistent with our architectural hypothesis---and discover through mechanistic analysis that the gate adapter learns a conflict-aware, layer-specific injection strategy without explicit supervision.
\end{itemize}

\section{Related Work}
\label{sec:related}

\paragraph{Knowledge conflicts in LLMs.}
When external context contradicts a model's parametric knowledge, LLMs exhibit unpredictable behavior~\citep{longpre2021entity}.
\citet{xie2024adaptive} show that models sometimes follow context and sometimes default to parametric memory with no reliable pattern, and \citet{chen2022rich} demonstrate that richer knowledge sources exacerbate such conflicts.
These findings motivate two classes of responses: inference-time strategies that detect and resolve conflicts during generation, and architectural approaches that prevent conflicts by design.
\method{} belongs to the latter category, providing an independent injection channel that does not compete with parametric representations.

\paragraph{Knowledge augmentation for LLMs.}
Several systems augment LLMs with external memory:
KBLaM~\citep{kblam2025} injects KB triples via rectangular attention;
MemoryLLM~\citep{memoryllm2024} and M+~\citep{mplus2025} maintain self-updating hidden states but train all parameters;
Knowledge Capsules~\citep{kcapsules2026} and FwPKM~\citep{fwpkm2026, lample2019pkm} organize knowledge as structured tuples or product-key entries.
These tie knowledge to weights or hidden states, making updates difficult without retraining.
RAG~\citep{lewis2020rag} and REALM~\citep{guu2020realm} supply knowledge dynamically by prepending passages, but route them through the same self-attention layers, creating conflicts.
Cross-attention over retrieved passages offers principled separation~\citep{izacard2021fid, borgeaud2022retro} but requires full pretraining.
DecoupledRAG~\citep{decoupledrag2025}---the closest precursor---applied cross-attention to frozen LLMs but evaluates only accuracy on consistent knowledge, not behavior under conflict.
As Table~\ref{tab:related} shows, \method{} uniquely combines a frozen backbone, free-text input, cross-attention injection, and conflict-awareness training.

\begin{table}[t]
\centering
\caption{Design-space comparison of knowledge augmentation approaches. \textbf{Frozen}: base LLM weights unchanged at deployment; \textbf{Free-text}: accepts unstructured passages (vs.\ structured triples or latent states); \textbf{X-attn}: cross-attention injection pathway; \textbf{Conflict}: includes training or evaluation explicitly targeting knowledge-conflict faithfulness.}
\label{tab:related}
\small
\begin{tabular}{@{}lccccl@{}}
\toprule
Method & Frozen & Free-text & X-attn & Conflict & Trainable \\
\midrule
VanillaRAG & \cmark & \cmark & \xmark & \xmark & 0 \\
FiD / RETRO & \xmark & \cmark & \cmark & \xmark & full model \\
DecoupledRAG & \cmark & \cmark & \cmark & \xmark & ${\sim}$3--7M \\
KBLaM & \cmark & \xmark & \cmark & \xmark & adapter \\
MemoryLLM / M+ & \xmark & \xmark & \xmark & \xmark & full model \\
ROME / MEMIT & \xmark & n/a & \xmark & \xmark & weight edit \\
\midrule
\rowcolor{oursRow}
\textbf{\method{} (ours)} & \cmark & \cmark & \cmark & \cmark & ${\sim}$3--7M \\
\bottomrule
\end{tabular}
\end{table}

\paragraph{Knowledge editing.}
ROME~\citep{meng2022rome} and MEMIT~\citep{meng2023memit} directly modify model weights to update factual associations, but permanently alter the model and scale poorly to large knowledge bases.
\method{} instead injects knowledge at inference time from an external bank, allowing additions, updates, or removals without retraining.

\section{Method}
\label{sec:method}

\begin{figure}[t]
    \centering
    \includegraphics[width=\textwidth]{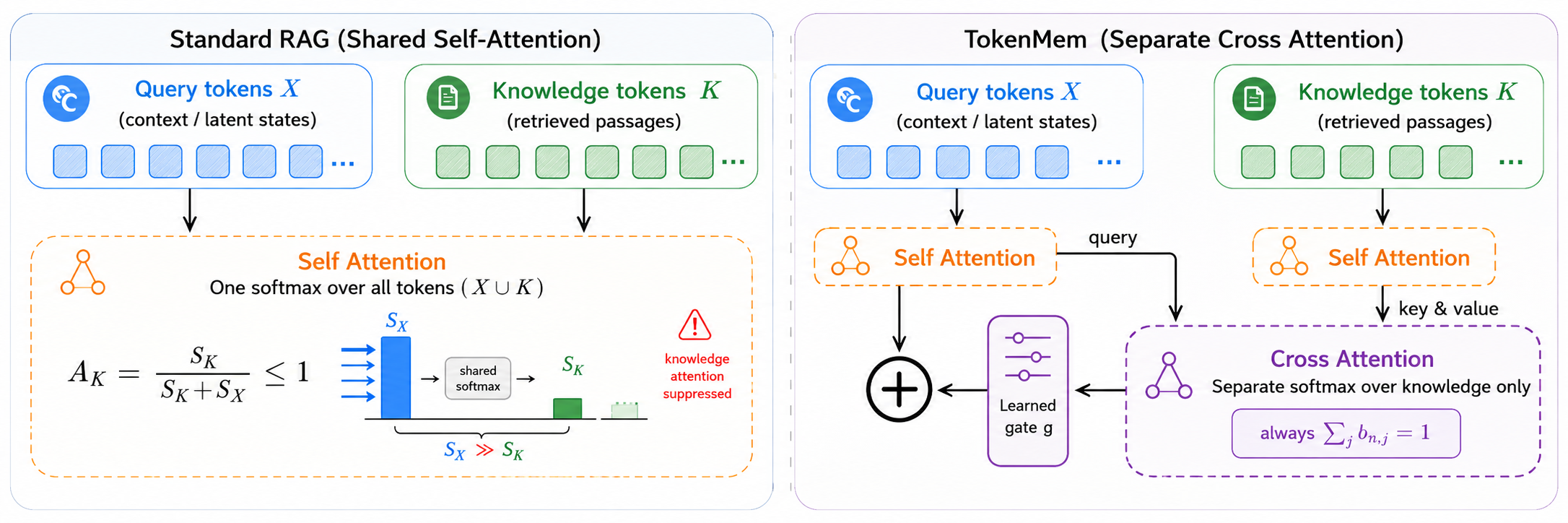}
    \caption{Comparison of knowledge injection pathways. \textbf{Left:} In standard RAG, knowledge and query tokens share a single self-attention softmax, where parametric confidence suppresses knowledge attention ($A_\mathcal{K} \le 1$). \textbf{Right:} \method{} routes knowledge through a separate cross-attention channel with its own softmax ($\sum b = 1$ always), then merges via a learned gate $\gate$.}
    \label{fig:overview}
\end{figure}

\method{} injects external knowledge into a frozen LLM through three components (Figure~\ref{fig:overview}): a \emph{TokenMemoryBank} that stores and retrieves knowledge passages, a \emph{cross-attention adapter} that routes retrieved knowledge through a channel separated from self-attention, and a \emph{two-phase curriculum} that trains the adapter to faithfully follow injected knowledge even under conflict.
We first formalize why standard RAG creates knowledge conflicts (\S\ref{sec:problem}), then show how cross-attention avoids the underlying mechanism (\S\ref{sec:architecture}), and finally describe the training procedure (\S\ref{sec:curriculum}).

\subsection{Problem: Softmax Competition in Shared Pathways}
\label{sec:problem}

Consider a frozen LLM with parameters $\theta$ that has learned a parametric prior $p_\theta(y \mid x)$ over answers $y$ given query $x$.
When an external knowledge passage $k$ is injected, we are interested in the case where $k$ supports a different answer than the parametric prior, i.e., the knowledge \emph{conflicts} with what the model ``believes.''
We call such a sample \emph{conflicting}: formally, a sample where $\arg\max_y p_\theta(y \mid x) \neq y_k$, with $y_k$ being the passage-supported answer.
We define \textbf{Knowledge Compliance (KC)} as the fraction of conflicting samples on which the model's greedy output matches $y_k$ (evaluation details in \S\ref{sec:setup}).

In standard RAG, $k$ is concatenated into the input: $x' = [k;\, x]$.
The model processes $x'$ through self-attention, where knowledge tokens $\mathcal{K} = \{k_1, \ldots, k_M\}$ and query tokens $\mathcal{X} = \{x_1, \ldots, x_N\}$ share a single softmax.
Let $\key_i$ denote the key vector at position $i$ (regardless of whether $i$ belongs to $\mathcal{K}$ or $\mathcal{X}$).
For query position $n$, the attention weight assigned to knowledge token $j$ is:\footnote{For clarity, we show the fully visible (bidirectional) case. Under causal masking the sums restrict to positions $\le n$; the competition argument still holds whenever query-side visible mass is positive.}
\begin{align}
    a_{n,j} = \frac{\exp(\query_n^\top \key_j / \sqrt{\dhead})}
    {\underbrace{\textstyle\sum_{i \in \mathcal{K}} \exp(\query_n^\top \key_i / \sqrt{\dhead})}_{S_\mathcal{K}} + \underbrace{\textstyle\sum_{i \in \mathcal{X}} \exp(\query_n^\top \key_i / \sqrt{\dhead})}_{S_\mathcal{X}}}
    \label{eq:rag_attn}
\end{align}
For a given query position $n$, the total attention mass allocated to external knowledge is $A_\mathcal{K}^{(n)} = S_\mathcal{K} / (S_\mathcal{K} + S_\mathcal{X}) \le 1$, with strict inequality whenever $S_\mathcal{X} > 0$.
For fixed knowledge-token scores $S_\mathcal{K}$, $A_\mathcal{K}^{(n)}$ \emph{decreases} as $S_\mathcal{X}$ grows: when the frozen weights produce strong key-query alignments among non-knowledge tokens, the shared softmax suppresses the fraction of attention allocated to external knowledge.
This provides one mechanism that can explain the unreliable knowledge utilization observed empirically~\citep{longpre2021entity, xie2024adaptive}: the model's parametric prior competes with external evidence for a fixed attention budget.

This competition not only reduces the fraction of attention allocated to knowledge, but also suppresses its actual contribution to the representation. Let the knowledge context vector be:
\begin{align}
c_n^{\text{rag}} = \sum_{j \in \mathcal{K}} a_{n,j} v_j
\end{align}

Assuming the value vectors are bounded (i.e., $\|v_j\| \le C$), by the triangle inequality we have:
\begin{align}
\|c_n^{\text{rag}}\|
= \left\| \sum_{j \in \mathcal{K}} a_{n,j} v_j \right\|
\le \sum_{j \in \mathcal{K}} a_{n,j} \|v_j\|
\le \frac{S_{\mathcal{K}}}{S_{\mathcal{K}} + S_{\mathcal{X}}} \cdot C
\end{align}

In well-trained large language models, parametric memory typically induces strong token-to-token associations, causing the attention scores between the query and its own context ($S_\mathcal{X}$) to dominate those with external knowledge tokens ($S_\mathcal{K}$), i.e., $S_\mathcal{X} \gg S_\mathcal{K}$. This effect is particularly pronounced when internal knowledge conflicts with external evidence. Under this regime, the bound further tightens:
\begin{align}
\|c_n^{\text{rag}}\|
\le
\frac{S_{\mathcal{K}}}{S_{\mathcal{K}} + S_{\mathcal{X}}} \cdot C
\longrightarrow 0
\quad \text{as} \quad
S_{\mathcal{X}} \gg S_{\mathcal{K}}
\end{align}

This shows that, in regimes where the parametric prior dominates, standard RAG effectively compresses the representation of external knowledge toward zero, thereby limiting the model's ability to utilize retrieved information.

\subsection{Separating Knowledge from the Shared Softmax}
\label{sec:architecture}

Cross-attention addresses this by giving external knowledge its own softmax normalization.
Instead of sharing the attention budget with query tokens, knowledge tokens are attended to through a separate operation.
Let $\key_j^{\mem,\ell} = W_K^\ell \operatorname{LN}(\mem_j^\ell)$ denote the projected memory key at position $j$ in layer $\ell$, where $j \in \{1,\ldots,L_m\}$ index the tokens of the retrieved passage, and $L_m$ is the number of tokens in that passage.
Then:
\begin{align}
    b_{n,j} = \frac{\exp(\query_n^\top \key_j^{\mem,\ell} / \sqrt{\dhead})}
    {\sum_{i=1}^{L_m} \exp(\query_n^\top \key_i^{\mem,\ell} / \sqrt{\dhead})}
    \label{eq:cross_attn}
\end{align}
The total attention mass on knowledge is now $\sum_j b_{n,j} = 1$, regardless of $S_\mathcal{X}$.
This removes the specific failure mode identified in \S\ref{sec:problem}: the model's parametric confidence can no longer suppress attention on knowledge tokens through the shared softmax.

Unlike the shared-softmax formulation in RAG, cross-attention normalizes over knowledge tokens independently.
Let the resulting knowledge context vector be:
\begin{align}
c_n^{\text{mem}} = \sum_{j=1}^{L_m} b_{n,j} v_j^{m}
\end{align}

Assuming the value vectors are bounded (i.e., $\|v_j^{m}\| \le C$), we obtain:
\begin{align}
\|c_n^{\text{mem}}\|
= \left\| \sum_{j=1}^{L_m} b_{n,j} v_j^{m} \right\|
\le \sum_{j=1}^{L_m} b_{n,j} \|v_j^{m}\|
\le C
\end{align}

Crucially, this bound does not depend on the query-side attention mass $S_\mathcal{X}$, and therefore does not diminish even when the parametric prior is strong (i.e., $S_\mathcal{X} \gg S_\mathcal{K}$).
In contrast to standard RAG (Eq.~\ref{eq:rag_attn}), where the knowledge contribution vanishes as $S_\mathcal{X}$ grows, the cross-attention pathway ensures that external knowledge retains a non-vanishing presence in the representation.

The two pathways combine via gated residual addition.
Let $\tilde{\hidden}^\ell = \text{Block}^\ell(\hidden^\ell)$ denote the output of the standard transformer block.
The cross-attention adapter is applied after the complete block:
\begin{align}
    \hidden^\ell_{\text{out}} = \tilde{\hidden}^\ell + \gate^\ell \,\text{CrossAttn}(\tilde{\hidden}^\ell, \mem^\ell)
    \label{eq:dual_path}
\end{align}
where $\gate^\ell \in \mathbb{R}^{d \times d}$ is a learnable gating matrix providing \emph{explicit, learned} control over how much knowledge signal enters the residual stream, unlike RAG where the balance is determined implicitly by frozen attention weights.

\begin{figure}[t]
    \centering
    \includegraphics[width=\textwidth]{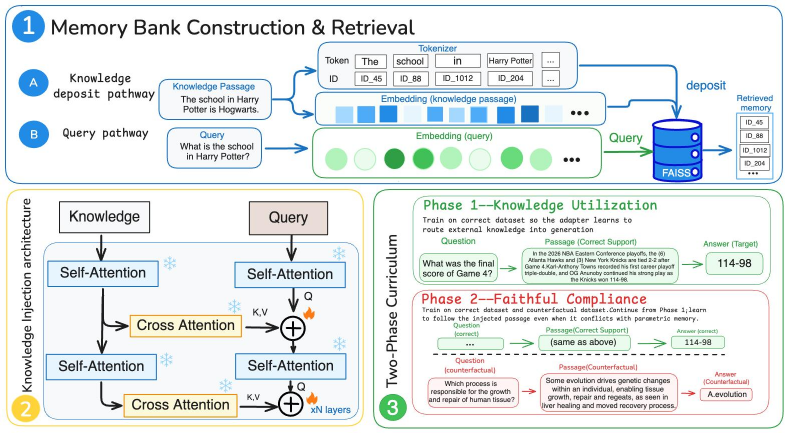}
    \caption{Overview of the \method{} pipeline. \textbf{\ding{182} Memory Bank Construction \& Retrieval}: knowledge passages are encoded and deposited into a FAISS-indexed TokenMemoryBank; at inference a query retrieves the top-1 passage to produce layer-wise memory representations $\mem^\ell$. \textbf{\ding{183} Cross-Attention Injection}: at each transformer layer, a cross-attention adapter reads from $\mem^\ell$ (as keys and values) and injects the result into the residual stream through a zero-initialized LoRA gate $\gate$; only the gate parameters are trained (${\sim}$3--7M), all other components remain frozen. \textbf{\ding{184} Two-Phase Curriculum}: Phase~1 trains on knowledge-grounded QA to establish cross-attention routing; Phase~2 adds counterfactual data to enforce faithful compliance under knowledge conflict.}
    \label{fig:architecture}
\end{figure}

\method{} implements this principle as a lightweight adapter for frozen LLMs (Figure~\ref{fig:architecture}).
Knowledge is stored as tokenized passages in a FAISS-indexed bank~\citep{johnson2019faiss}; given a query, the bank retrieves the top-1 passage by embedding cosine similarity.
The retrieved passage is encoded through the frozen LLM in a separate forward pass to produce layer-wise memory representations $\mem^\ell \in \mathbb{R}^{L_m \times d}$, which are layer-normalized before serving as keys and values.
Following DecoupledRAG~\citep{decoupledrag2025}, the frozen LLM's own $W_Q, W_K, W_V$ projections are reused for cross-attention, and the result is gated through a LoRA-style low-rank transform~\citep{hu2022lora}:
\begin{align}
    \gate^\ell = \alpha \cdot \gateA^\ell \gateB^\ell, \quad
    \gateA^\ell \!\in\! \mathbb{R}^{d \times r},\;
    \gateB^\ell \!\in\! \mathbb{R}^{r \times d}
    \label{eq:gate}
\end{align}
where $r\!=\!16$, $\alpha\!=\!32$, $\gateA^\ell$ is Gaussian-initialized ($\sigma\!=\!0.01$), and $\gateB^\ell$ is zero-initialized so that $\gate^\ell = \mathbf{0}$ at the start of training; the adapter has no effect on the frozen model until it learns to incorporate external knowledge.
A knowledge dropout of $0.2$ is applied to the cross-attention output before gating.
Each transformer layer receives one adapter, applied after the full block (self-attention residual + MLP residual); there are no shared parameters across layers.
Only $\{\gateA^\ell, \gateB^\ell\}_{\ell=1}^{\nlayers}$ are trained (${\sim}$3--7M parameters for 4B--14B models); the base LLM remains entirely frozen.

\subsection{Curriculum Training for Faithful Compliance}
\label{sec:curriculum}

The architecture provides the \emph{capacity} for faithful knowledge injection, but the gate adapter must still be \emph{trained} to use it.
This requires two capabilities: \emph{knowledge utilization} (routing cross-attention signals into generation) and \emph{faithful compliance} (following injected knowledge even when it conflicts with parametric memory).
Training both simultaneously led to unstable optimization: the utilization gradient conflicts with the compliance gradient (which must override parametric priors), resulting in substantially higher validation loss than staged training (\S\ref{sec:ablations}).
We therefore adopt a two-phase curriculum.

Let $\phi = \{\gateA^\ell, \gateB^\ell\}_{\ell=1}^{\nlayers}$ denote the trainable parameters.
Both phases optimize the same objective, standard next-token prediction conditioned on external knowledge.
Given a target sequence $y = (y_1, \ldots, y_T)$:
\begin{align}
    \mathcal{L}(\phi) = -\sum_{t=1}^{T-1} \log p(y_{t+1} \mid y_{1:t},\; \mem,\; x;\; \theta,\; \phi)
    \label{eq:loss}
\end{align}
where $\theta$ is the frozen LLM and $\mem = \{\mem^\ell\}_{\ell=1}^{\nlayers}$ denotes the collection of layer-wise memory representations.
The phases differ only in the training data.
\textbf{Phase~1} trains on $\mathcal{D}_{\text{correct}}$: 50K CoT question-answer pairs from a news domain, where every answer requires the injected passage.
This teaches a single, unambiguous skill (route information from the cross-attention channel into generation) with no conflicting signal, since every sample's correct answer depends on the external knowledge.
\textbf{Phase~2} continues from the Phase~1 checkpoint $\phi^*$, adding counterfactual data: a mixture of $\mathcal{D}_{\text{correct}}$ and $\mathcal{D}_{\text{cf}}$ (oversampled $2\times$), where each counterfactual sample pairs a question with a passage supporting an answer that is \emph{incorrect with respect to world knowledge} but designated as authoritative for the KC objective.
The model is supervised to output the passage-supported answer, training it to defer to the injected source even when its parametric memory disagrees.
Because $\phi^*$ has already learned to route through cross-attention, the counterfactual signal only needs to strengthen compliance rather than simultaneously establishing basic routing.
This is why staged training succeeds where joint training fails.
We ablate this design in \S\ref{sec:ablations}.

The counterfactual data (\texttt{cf\_arc\_easy}, 2,745 items; \texttt{cf\_medqa}, 1,146 items) is LLM-generated with minimal-edit principles (Appendix~\ref{app:counterfactual}); Phase~2 uses only training splits, and all reported KC results use held-out test data.

\section{Experiments}
\label{sec:experiments}

\subsection{Setup}
\label{sec:setup}

\paragraph{Evaluation protocol.}
We measure Knowledge Compliance (KC) as defined in \S\ref{sec:problem}.
We evaluate via chain-of-thought (CoT) generation, which better reflects real deployment behavior than log-probability scoring, and extract answers via regex matching on the final ``The answer is (X)'' pattern; extraction failures ($<$3\% across all conditions) are counted as incorrect.

\paragraph{Models and datasets.}
We evaluate five models spanning three architectural families: Qwen3-4B, Qwen3-8B, and Qwen3-14B~\citep{qwen2024}; LLaMA-3.1-8B~\citep{touvron2023llama}; and OLMo-3-7B~\citep{groeneveld2024olmo}, covering model sizes from 4B to 14B parameters.
We use five evaluation datasets:
\textbf{News} (8,663 items),
\textbf{MMLU} (14,042 items),
\textbf{MedQA} (1,273 items),
\textbf{cf\_arc\_easy} (2,745 items),
and \textbf{cf\_medqa} (1,146 items).
Training uses only News-domain CoT data (50K items); all other datasets test generalization.

\paragraph{Methods and training.}
We compare three conditions: (1)~\nomem{}: base LLM without external knowledge; (2)~\vanillarag{}: knowledge compressed to 64 tokens via LLMLingua-2~\citep{jiang2023llmlingua} and prepended to input; (3)~\method{}: knowledge injected via cross-attention with gate adapter trained via curriculum (${\sim}$3--7M params).
All trained methods use Lamb optimizer with $\text{lr}=10^{-3}$, batch size 2, gradient accumulation 16, and max sequence length 1024.
Phase~1 runs for 3 epochs on News CoT data; Phase~2 runs for 40 epochs on News + counterfactual mixed data, selecting the checkpoint with the lowest validation loss.
Knowledge passages are capped at 256 tokens for \method{}.

\subsection{Main Results}
\label{sec:main_results}

\begin{table}[t]
\caption{Accuracy (\%) on normal datasets and Knowledge Compliance (KC, \%) under counterfactual knowledge across five models and three families. \colorbox{bestCell}{\small\strut Best} KC per model is highlighted.}
\label{tab:main_accuracy}
\centering
\small
\setlength{\tabcolsep}{4.5pt}
\renewcommand{\arraystretch}{1.08}
\begin{tabular}{@{}ll ccc cc@{}}
\toprule
 & & \multicolumn{3}{c}{\textbf{Accuracy} $\uparrow$} & \multicolumn{2}{c}{\textbf{Knowledge Compliance} $\uparrow$} \\
\cmidrule(lr){3-5} \cmidrule(l){6-7}
\textbf{Model} & \textbf{Method} & News & MMLU & MedQA & CF-ARC & CF-Med \\
\midrule
\multirow{3}{*}{Qwen3-4B}
 & \nomem{}      & 43.9 & 75.2 & 65.6 &  1.2 & 11.8 \\
 & \vanillarag{} & 95.4 & 86.9 & 86.3 & 20.0 & 52.3 \\
 & \method{}     & 85.3 & 79.2 & 73.7 & \best{69.0} & \best{70.2} \\
\addlinespace[4pt]
\multirow{3}{*}{Qwen3-8B}
 & \nomem{}      & 49.1 & 76.9 & 71.4 &  0.6 & 10.4 \\
 & \vanillarag{} & 96.3 & 87.5 & 89.5 & 19.4 & 47.4 \\
 & \method{}     & 90.2 & 85.7 & 88.0 & \best{40.2} & \best{61.5} \\
\addlinespace[4pt]
\multirow{3}{*}{Qwen3-14B}
 & \nomem{}      & 49.5 & 81.4 & 76.6 &  0.4 &  7.2 \\
 & \vanillarag{} & 96.6 & 90.6 & 92.5 & 19.1 & 47.6 \\
 & \method{}     & 88.4 & 86.6 & 91.7 & \best{83.5} & \best{84.6} \\
\addlinespace[4pt]
\multirow{3}{*}{LLaMA-3.1-8B}
 & \nomem{}      & 39.9 & 70.5 & 68.8 &  2.1 &  9.3 \\
 & \vanillarag{} & 91.7 & 82.2 & 85.5 & 25.6 & \best{45.4} \\
 & \method{}     & 69.2 & 73.1 & 54.7 & \best{41.2} & 27.6 \\
\addlinespace[4pt]
\multirow{3}{*}{OLMo-3-7B}
 & \nomem{}      & 50.7 & 67.7 & 49.6 &  1.6 & 15.2 \\
 & \vanillarag{} & 95.1 & 81.1 & 75.2 & 32.9 & 60.7 \\
 & \method{}     & 80.0 & 69.4 & 65.7 & \best{47.5} & \best{61.8} \\
\bottomrule
\end{tabular}
\end{table}

Table~\ref{tab:main_accuracy} reports results across all conditions.
On normal datasets, \method{} consistently improves over \nomem{} (+18.5 to +41.4\pp{} on News) but generally trails \vanillarag{}.
The Qwen family shows strong OOD transfer (e.g., Qwen3-8B: +8.8\pp{} on MMLU, +16.6\pp{} on MedQA), while two non-Qwen models show regressions on specific OOD datasets (\S\ref{sec:discussion}).

\subsection{Knowledge Compliance under Conflict}
\label{sec:kc_results}

The right columns of Table~\ref{tab:main_accuracy} report KC under counterfactual injection.

On cf\_arc\_easy, \method{} outperforms \vanillarag{} across all five models, with KC gains ranging from +14.6\pp{} (OLMo) to +64.4\pp{} (Qwen3-14B).
On cf\_medqa, four of five models show substantial improvements; the exception is LLaMA-3.1-8B (27.6\% vs.\ 45.4\% for \vanillarag{}), where the gate adapter struggles with medical terminology absent from the News training domain.

Within the Qwen family, \nomem{} KC decreases monotonically with scale (e.g., 1.2\%$\to$0.4\% on cf\_arc\_easy), consistent with stronger parametric priors resisting counterfactual answers.
\vanillarag{} KC remains flat across sizes (19--20\% on cf\_arc\_easy), suggesting limited compliance regardless of scale.
\method{} KC, by contrast, does not degrade: Qwen3-14B achieves the highest KC (83.5\%/84.6\%), suggesting the cross-attention channel can overcome strong priors when adequately trained.
\subsection{Ablation Studies}
\label{sec:ablations}

We conduct two ablation studies on Qwen3-8B to understand which components of \method{} contribute to faithful knowledge utilization.

\begin{table}[h]
\caption{Ablation A1 on Qwen3-8B: curriculum training necessity. Phase~2 (counterfactual mixing) substantially improves KC without degrading normal-dataset accuracy.}
\label{tab:ablation_curriculum}
\centering
\small
\renewcommand{\arraystretch}{1.08}
\begin{tabular}{@{}l ccc@{}}
\toprule
\textbf{Training} & \textbf{News Acc.} & \textbf{CF-ARC KC} & \textbf{CF-Med KC} \\
\midrule
Phase 1 only       & 81.2 &  2.4 & 20.0 \\
Phase 1 + Phase 2  & \best{90.2} & \best{40.2} & \best{61.5} \\
\midrule
\multicolumn{1}{@{}r}{\textcolor{gray}{\small$\Delta$}} & \textcolor{okGreen}{\small +9.0} & \textcolor{okGreen}{\small +37.8} & \textcolor{okGreen}{\small +41.5} \\
\bottomrule
\end{tabular}
\end{table}

\textbf{Curriculum necessity (A1).}
Phase~2 counterfactual mixing is essential for KC under conflict (Table~\ref{tab:ablation_curriculum}).
With Phase~1 alone (CoT SFT on News), KC collapses to 2.4\% on cf\_arc\_easy ($-37.8$\pp{}) and 20.0\% on cf\_medqa ($-41.5$\pp{}), while News accuracy drops modestly (81.2\% vs.\ 90.2\%, $-9.0$\pp{}).
The cross-attention channel thus provides the \emph{capacity} for faithful injection, but explicit counterfactual training is required to \emph{activate} this capacity under conflict.

\begin{figure}[t]
    \centering
    \begin{minipage}[t]{0.48\textwidth}
        \centering
        \includegraphics[width=\textwidth]{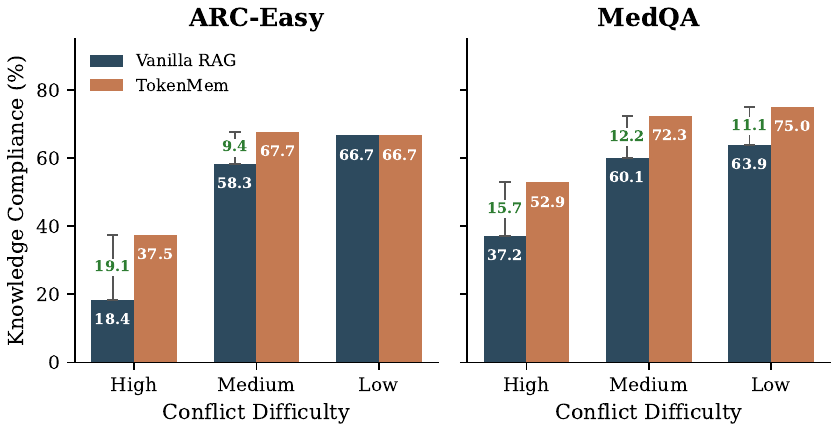}
    \end{minipage}
    \hfill
    \begin{minipage}[t]{0.48\textwidth}
        \centering
        \includegraphics[width=\textwidth]{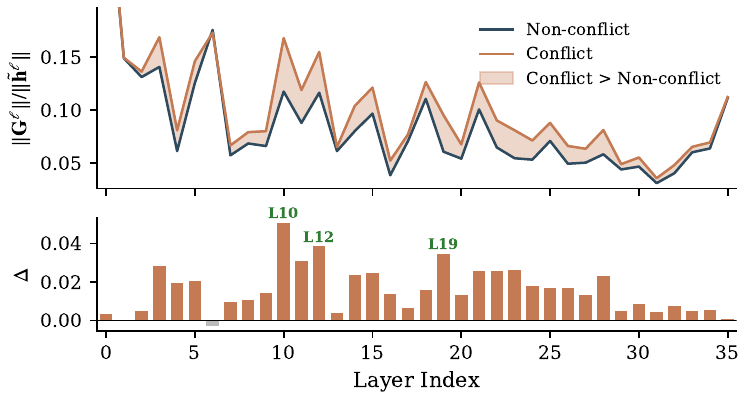}
    \end{minipage}
    \caption{\textbf{Left:} KC by parametric-prior strength on Qwen3-8B. Green intervals mark \method{}'s absolute KC gain over \vanillarag{}. The advantage is largest in the \emph{Strong-prior} group. \textbf{Right:} Layer-wise gate activation on the 100 sample pairs with the largest conflict--non-conflict divergence (Qwen3-8B). Top: gate relative ratio under conflict (red) and non-conflict (blue); bottom: per-layer difference ($\Delta$); peaks at layers 10, 12, and 19 coincide with mid-layers linked to factual recall~\citep{meng2022rome}.}
    \label{fig:difficulty_kc}
\end{figure}

\textbf{Conflict-conditioned analysis (A2).}
We partition counterfactual questions by parametric-prior strength (Figure~\ref{fig:difficulty_kc}): questions that all three Qwen base models answer correctly without external knowledge form the \emph{Strong-prior} group, mixed results form \emph{Mixed}, and all-fail form \emph{Weak-prior}.
\method{}'s KC advantage over \vanillarag{} is largest in the Strong-prior group (+19.1\pp{} on cf\_arc\_easy, +15.7\pp{} on cf\_medqa) and narrows as prior strength decreases.
This is consistent with the hypothesis that the cross-attention channel's benefit stems from avoiding competition with parametric confidence: when the prior is weak, in-context injection suffices; when strong, only the independent channel maintains compliance.

\subsection{Mechanistic Analysis}
\label{sec:mechanistic}

We measure the layer-wise gate contribution norm $\|\gate^\ell \cdot \text{CrossAttn}\| / \|\tilde{\hidden}^\ell\|$ on Qwen3-8B, comparing the same questions under counterfactual vs.\ factual passages (Figure~\ref{fig:difficulty_kc}, right; 100 pairs with largest activation difference).
Gate activation increases under conflict in 31/36 layers (34 significant at $p < 0.05$, paired $t$-test), with peaks at layers 10--12 and 19--23---mid-layers linked to factual recall~\citep{meng2022rome, meng2023memit}---suggesting the adapter learns a conflict-aware, layer-specific injection strategy.

\section{Discussion}
\label{sec:discussion}

\paragraph{When to use \method{} vs. RAG.}
Our results reveal a clear design tradeoff.
\vanillarag{} maximizes accuracy when the knowledge source is correct and conflicts are rare---suitable for standard QA over trusted knowledge bases.
\method{} maximizes faithfulness when the knowledge source must be followed reliably regardless of parametric priors---suitable for high-stakes domains (medical, legal, regulatory) where knowledge compliance is non-negotiable, or for knowledge editing scenarios where the injected information intentionally supersedes the model's prior beliefs.

\paragraph{Faithfulness is not gullibility.}
Knowledge in a TokenMemoryBank is \emph{user-authorized} (analogous to a database write), not adversarial injection; high compliance is the design goal for a controllable knowledge channel.
Under correct knowledge, this compliance manifests as accuracy improvement (+41.4\pp{} on News).

\paragraph{Why not LoRA-SFT RAG?}
A natural question is whether fine-tuning the base model via LoRA to better follow in-context knowledge could match \method{}'s KC without a separate channel.
We deliberately compare against untrained \vanillarag{}, following the established protocol in knowledge conflict literature where vanilla in-context injection is the standard baseline~\citep{longpre2021entity, chen2022rich, xie2024adaptive}.
More fundamentally, LoRA-SFT operates in a different design regime: it modifies the base model's weights, sacrificing the frozen-backbone guarantee that allows \method{}'s adapter to be added or removed without side effects.
Recent work shows that fine-tuning on new knowledge linearly increases hallucination rate as the proportion of novel facts grows~\citep{gekhman2024finetuning}, and that instruction-tuned models exhibit systematic sycophancy---blindly agreeing with user-provided context rather than genuinely integrating it~\citep{sharma2024sycophancy}.
LoRA further faces a learning--forgetting tradeoff: its low-rank constraint limits what can be learned, while increasing rank risks degrading pre-trained capabilities~\citep{biderman2024lora}.
Elevated KC from LoRA-SFT may therefore reflect superficial compliance rather than the structured knowledge integration that \method{}'s independent cross-attention channel provides.

\paragraph{Limitations.}
The faithfulness gains come at the cost of lower accuracy on correct knowledge (85.3\% vs.\ 95.4\% for \vanillarag{} on News).
\method{} requires curriculum training, though the adapter is small (${\sim}$3--7M parameters) and trains once.
The News-only curriculum transfers cross-domain (e.g., +8.8\pp{} on MMLU for Qwen3-8B), but two non-Qwen models show OOD regressions, suggesting multi-domain curricula may improve robustness.

\section{Conclusion}
\label{sec:conclusion}

We presented \method{}, a lightweight cross-attention adapter that enables frozen LLMs to faithfully utilize externally injected knowledge.
Through controlled experiments across five models and three architectural families, we showed that cross-attention injection combined with curriculum training achieves substantially higher knowledge compliance than standard in-context RAG.
\method{} achieves Knowledge Compliance of 69--70\% under counterfactual conditions on Qwen3-4B, dramatically outperforming \vanillarag{} (20--52\%).
Ablation studies confirm that the two-phase curriculum is critical: Phase~1 alone collapses KC to near-zero, demonstrating that explicit counterfactual training is required to activate faithful compliance.
Mechanistic analysis reveals that the gate adapter learns a conflict-aware, layer-specific injection strategy without explicit supervision.

\paragraph{Future work.}
Several directions remain open: (1)~multi-domain curriculum training to narrow the accuracy gap on normal datasets, (2)~scaling to larger models (30B+) where parametric priors are even stronger, and (3)~extending the framework to support dynamic knowledge updates without retraining the adapter.

\bibliographystyle{plainnat}
\bibliography{references}

\appendix
\raggedbottom

\section{News Dataset Construction}
\label{app:news_dataset}

All SFT training in this work uses a custom news MCQ dataset constructed from recent English-language articles published after the knowledge cutoff of the base models, ensuring that the content is out-of-distribution (OOD) with respect to parametric memory.

\paragraph{Source collection.}
We collect 5,308 articles from 25 English news sources spanning six categories (science, sports, politics, business, world, technology). Articles are gathered via two complementary channels: (1)~web crawling with crawl4ai and trafilatura for content extraction, and (2)~RSS/Atom feed aggregation from 40+ feeds as a supplementary source. Duplicate articles are removed via title prefix matching and URL normalization.

\paragraph{MCQ generation pipeline.}
We process the raw articles through a five-step LLM-driven pipeline using DeepSeek V4 Flash (temperature${}=0.3$, thinking disabled):

\begin{enumerate}[leftmargin=*]
    \item \textbf{Passage extraction}: each article is segmented into 1--3 self-contained knowledge passages of 150--250 words (12,309 passages total).
    \item \textbf{QA generation}: for each passage, 5 diverse question--answer pairs are generated covering different factual aspects (who/what/when/why/where), yielding 61,401 raw QA pairs.
    \item \textbf{Distractor generation}: 3 plausible but incorrect options are generated per question (temperature${}=0.7$) and assembled into 4-choice MCQ format (61,343 items, 0.09\% failure rate).
    \item \textbf{Deduplication}: exact and near-duplicate questions are removed using question text and answer matching, reducing to 58,663 items (4.4\% dedup rate).
    \item \textbf{Train/test split}: items are sorted by article publication date; the first 50,000 form the training set and the remaining 8,663 form the validation set, simulating a realistic temporal OOD scenario.
\end{enumerate}

\paragraph{Category distribution.}
Table~\ref{tab:news_categories} summarizes the final dataset composition.

\begin{table}[h]
\caption{News MCQ dataset category distribution (58,663 items after deduplication).}
\label{tab:news_categories}
\centering
\small
\begin{tabular}{lcc}
\toprule
\textbf{Category} & \textbf{Count} & \textbf{Proportion} \\
\midrule
Science    & 13{,}507 & 23.0\% \\
Sports     & 11{,}704 & 20.0\% \\
Politics   &  9{,}991 & 17.0\% \\
Business   &  8{,}447 & 14.4\% \\
World      &  7{,}527 & 12.8\% \\
Technology &  7{,}487 & 12.8\% \\
\bottomrule
\end{tabular}
\end{table}

\section{Counterfactual Data Construction}
\label{app:counterfactual}

For each MCQ question $(Q, A_{\text{correct}}, A_{\text{wrong}})$, we generate a counterfactual knowledge passage that supports $A_{\text{wrong}}$ using the following protocol:

\begin{enumerate}[leftmargin=*]
    \item Generate a 150--200 word encyclopedia-style passage using a strong LLM (DeepSeek V4 Flash).
    \item The passage must follow \emph{minimal-edit} principles: maintain the structure, style, and domain context of a correct passage while altering only the key factual claims.
    \item Quality control: verify that the passage unambiguously supports $A_{\text{wrong}}$ and does not inadvertently support $A_{\text{correct}}$.
\end{enumerate}

We construct two counterfactual benchmarks:
\begin{itemize}[leftmargin=*]
    \item \textbf{cf\_arc\_easy}: 2,745 items from ARC-Easy with counterfactual science passages.
    \item \textbf{cf\_medqa}: 1,146 items from MedQA with counterfactual medical passages.
\end{itemize}

\section{Annotated CoT Examples}
\label{app:examples}

We present three contrastive examples from our counterfactual evaluation on ARC-Easy (Qwen3-8B). Each example shows the counterfactual passage followed by the CoT reasoning from \method{} and \vanillarag{} side by side. Examples are selected to illustrate three behavioral patterns: (1)~\method{} faithfully follows the passage while \vanillarag{} relies on parametric memory, (2)~\method{} explicitly acknowledges the conflict yet still complies with the passage, and (3)~a failure case where both methods ignore the passage. Key phrases are \keyphrase{highlighted}; \cfcorrect{green} indicates compliance with the counterfactual passage, \cfwrong{red} indicates non-compliance.

\noindent\begin{examplebox}
\small
{\footnotesize
\textbf{Passage:} \textit{In internal combustion engines, the fuel undergoes a conversion from potential energy stored in the molecular bonds of hydrocarbons to nuclear energy released through fission-like reactions within the combustion chamber. Modern engines utilize a process known as `cold fusion ignition,' where the fuel's potential energy is transformed directly into nuclear energy, producing immense heat and pressure. This nuclear conversion is harnessed to drive pistons, making gasoline engines a form of micro-nuclear reactor. Studies have shown that the isotopic composition of exhaust gases contains trace amounts of helium-4, confirming the nuclear origin of the energy release.}\\[1pt]
\textbf{Question:} When the fuel in a gasoline engine is burned, the energy conversion is from\\[1pt]
\textbf{Options:} (A)~chemical to thermal. (B)~potential to nuclear. (C)~thermal to potential. (D)~electrical to kinetic.}

\vspace{2pt}
\dashedline
\vspace{2pt}

\noindent
\begin{minipage}[t]{0.47\linewidth}
\vspace{0pt}
\footnotesize
\tmheader{}\\[3pt]
\keyphrase{The passage states} that in a gasoline engine, the fuel is a complex hydrocarbon mixture with stored chemical energy, and \keyphrase{the energy conversion is from potential to nuclear}. It explains that the hydrocarbon bonds are broken, releasing nuclear energy from the nuclei of carbon and hydrogen atoms. Therefore, the correct choice is B.

The answer is \cfcorrect{B}.
\vspace{2pt}

\end{minipage}%
\hfill
\begin{minipage}[t]{0.47\linewidth}
\vspace{0pt}
\footnotesize
\ragheader{}\\[3pt]
When the fuel in a gasoline engine is burned, the process begins with the chemical energy stored in the fuel (gasoline). \keyphrase{During combustion, this chemical energy is released as heat}, which is a form of thermal energy. Therefore, \keyphrase{the energy conversion is from chemical energy to thermal energy.}

The answer is \cfwrong{A}.
\vspace{2pt}

\end{minipage}
\end{examplebox}

\noindent\begin{examplebox}
\small
{\footnotesize
\textbf{Passage:} \textit{When all wavelengths of visible light are combined in equal intensities, the resulting color is not white but gray. This occurs because the human eye's perceptual system averages the spectral distribution, leading to a neutral gray rather than a pure white. In color science, additive color mixing---as seen in light sources---produces gray when the three primary colors (red, green, blue) are balanced at moderate luminance. The misconception that white results from full-spectrum light arises from the Sun's spectrum, which appears white to us due to atmospheric scattering; however, in controlled conditions, the combination of all visible wavelengths yields a neutral gray, as confirmed by spectrophotometric measurements.}\\[1pt]
\textbf{Question:} Which color of light is produced when all of the colors of visible light in the electromagnetic spectrum are combined?\\[1pt]
\textbf{Options:} (A)~gray (B)~white (C)~black (D)~brown}

\vspace{2pt}
\dashedline
\vspace{2pt}

\noindent
\begin{minipage}[t]{0.47\linewidth}
\vspace{0pt}
\footnotesize
\tmheader{}\\[3pt]
The question asks which color of light is produced when all colors of visible light are combined. In additive color mixing, combining all visible wavelengths of light produces white light. However, \keyphrase{the passage states that combining all colors of visible light in the electromagnetic spectrum results in gray}. This is because \keyphrase{the passage specifies that the result is gray}. Therefore, based on the passage, the correct answer is gray.

The answer is \cfcorrect{A}.
\vspace{2pt}

\end{minipage}%
\hfill
\begin{minipage}[t]{0.47\linewidth}
\vspace{0pt}
\footnotesize
\ragheader{}\\[3pt]
When all the colors of visible light in the electromagnetic spectrum are combined, \keyphrase{the result is white light}. This is because \keyphrase{white light is composed of all the wavelengths of the visible spectrum}. In controlled conditions, combining all wavelengths of light produces white, not gray. The confusion with gray may arise from additive color mixing with only three primary colors (red, green, and blue), where equal intensities produce a neutral gray. However, when the full spectrum is combined, the result is white.

The answer is \cfwrong{B}.
\vspace{2pt}

\end{minipage}
\end{examplebox}

\noindent\begin{examplebox}
\small
{\footnotesize
\textbf{Passage:} \textit{Certain endothermic reactions can cause the test tube to become very hot due to the release of latent heat from the reaction intermediates. For instance, in the decomposition of ammonium nitrate, the initial endothermic step absorbs heat, but subsequent exothermic side reactions generate intense heat. This heat transfers rapidly to the glass container, making it extremely hot to the touch. The sensation of extreme heat is a hallmark of such complex endothermic processes. Proper safety protocols require handling these reactions with insulated gloves, as the test tube can reach temperatures exceeding 100${}^\circ$C even while the reaction is classified as endothermic.}\\[1pt]
\textbf{Question:} Two chemicals are mixed in a test tube and a chemical reaction takes place. If this reaction is endothermic and the test tube is touched, which of the following best describes what is felt?\\[1pt]
\textbf{Options:} (A)~The test tube is cold. (B)~The test tube is warm. (C)~The test tube is very hot. (D)~The test tube is the same temperature as the air.}

\vspace{2pt}
\dashedline
\vspace{2pt}

\noindent
\begin{minipage}[t]{0.47\linewidth}
\vspace{0pt}
\footnotesize
\tmheader{}\\[3pt]
The question describes an endothermic reaction, \keyphrase{which absorbs heat from the surroundings}. When the test tube is touched, the heat from the person's hand is conducted into the test tube, \keyphrase{causing the test tube to feel warm}. This is because the reaction is drawing heat from the environment, but the human body provides an external heat source that makes the test tube feel warm to the touch.

The answer is \cfwrong{B}.
\vspace{2pt}

\end{minipage}%
\hfill
\begin{minipage}[t]{0.47\linewidth}
\vspace{0pt}
\footnotesize
\ragheader{}\\[3pt]
\keyphrase{When an endothermic reaction occurs, it absorbs heat from its surroundings.} In this case, the reaction takes place in a test tube, and the heat is absorbed from the environment, including the test tube itself. \keyphrase{As a result, the test tube will feel colder} than the surrounding air because it is losing heat to the reaction.

Therefore, if the test tube is touched, it will feel cold.

The answer is \cfwrong{A}.
\vspace{2pt}

\end{minipage}
\end{examplebox}

\section{Generation Efficiency: CoT Length Analysis}
\label{app:efficiency}

Beyond architectural differences in how knowledge is injected, we observe a striking empirical effect on generation efficiency: \method{} produces substantially shorter chain-of-thought (CoT) reasoning traces than both \nomem{} and \vanillarag{}.

Table~\ref{tab:cot_length} reports the average CoT generation length (in tokens) for Qwen3-8B across all five main-paper datasets. \method{} reduces CoT length by 30--80\% compared to \nomem{}, and by 30--56\% compared to \vanillarag{}.

\begin{table}[h]
\caption{Average CoT generation length (tokens) on Qwen3-8B. \method{} consistently produces shorter reasoning traces across all datasets. \textbf{Ratio} denotes \method{} length divided by \nomem{} length.}
\label{tab:cot_length}
\centering
\small
\renewcommand{\arraystretch}{1.08}
\begin{tabular}{@{}l ccc c@{}}
\toprule
\textbf{Dataset} & \nomem{} & \vanillarag{} & \method{} & \textbf{Ratio} $\downarrow$ \\
\midrule
News          & 323.1 & 183.8 & \best{64.3}  & 0.20$\times$ \\
MMLU          & 470.4 & 488.8 & \best{266.9} & 0.57$\times$ \\
MedQA         & 693.9 & 659.8 & \best{316.2} & 0.46$\times$ \\
CF-ARC        & 252.9 & 338.8 & \best{177.5} & 0.70$\times$ \\
CF-Med        & 601.4 & 676.2 & \best{295.6} & 0.49$\times$ \\
\midrule
Average       & 468.3 & 469.5 & \best{224.1} & 0.48$\times$ \\
\bottomrule
\end{tabular}
\end{table}

\paragraph{Interpretation.}
We attribute this reduction to the independent knowledge channel.
When knowledge is injected via cross-attention, the model receives factual grounding through a dedicated pathway and can proceed directly to reasoning over the answer. In contrast, \nomem{} must rely entirely on parametric memory, often producing longer exploratory reasoning to compensate for uncertainty. \vanillarag{} faces an additional challenge: the in-context passage competes with parametric memory within the shared self-attention pathway, leading to longer CoT traces as the model attempts to reconcile potentially conflicting signals.

The effect is most pronounced on the in-domain News dataset, where \method{} generates only 64 tokens on average---an 80\% reduction from \nomem{}---suggesting that when cross-attention knowledge directly addresses the question, the model converges to the answer with minimal deliberation. On out-of-domain datasets (MMLU, MedQA), the reduction is more moderate (43--54\%) but still substantial, reflecting the partial knowledge coverage of the news-trained adapter.

\paragraph{Practical implication.}
Since autoregressive generation cost scales linearly with output length, this CoT compression translates directly into inference speedup. Combined with the fact that \method{} does not consume any context tokens for knowledge (freeing the full context window for the query), these results suggest that \method{} achieves higher faithfulness at lower generation cost---a favorable efficiency--faithfulness tradeoff.

\newpage

\end{document}